\documentclass[12pt]{article}

\usepackage{sbc-template}
\usepackage{graphicx,url}
\usepackage[utf8]{inputenc}
\usepackage[brazil]{babel}
\usepackage[tableposition=top]{caption}
\usepackage{svg}

\title{Time Series Prediction about Air Quality using LSTM-Based Models: A Systematic Mapping}

\author{Lucas L. S. Sachetti\inst{1}, Vinicius F. S. Mota\inst{1} }

\address{Computer Science Department -- Federal University of Espírito Santo
  (UFES)\\
  Mailbox 01-9011 -- 29060-970 -- Vitória -- ES -- Brazil
  \email{\{llopes,vinicius.mota\}@inf.ufes.br}
}

\begin{document} 

\maketitle

\begin{abstract}
  This systematic mapping study investigates the use of Long short-term memory networks to predict time series data about air quality, trying to understand the reasons, characteristics and methods available in the scientific literature, identify gaps in the researched area and potential approaches that can be exploited on later studies.
\end{abstract}


\section{Introduction}

Air pollution has become increasingly worrying for the population in recent times. As a result, data on air quality has been gradually increasing and the science underlying health-related impacts is also evolving rapidly \cite{who:hobook09}. There are several air pollutants, the most common of which are Carbon Monoxide (CO), Ozone (O$_{3}$), Nitrogen Dioxide (NO$_{2}$), Sulfur Dioxide (SO$_{2}$) and Particulate Material (PM$_{10}$ and PM$_{2.5}$), which are pollutants monitored to make the so-called Air Quality Index (AQI) which qualitatively says the state of the environment. The monitoring of these pollutants is usually done on a macro-scale by meteorological stations scattered at strategic points in cities. 

These pollutant data are said to be a time series, which is a set of observations ordered in time. This makes it possible to use analytical methods to tell whether the air quality is good for the population or not. Among these methods there is descriptive analysis, which only describes current and past events. Predictive analysis, which uses past data to try to predict future actions. There is also prescriptive analysis, which uses predictive and/or descriptive analysis to make or recommend decisions based on the results obtained by the analyzes.

There are some ways to make data prediction, including the use of artificial neural networks (ANN) that are commonly used to recognize patterns \cite{crone2010feature}. As the data have a strong relationship with time, one of the most recommended networks is a recurrent neural network (RNN) \cite{bone2003boosting} which uses previous information to serve as an understanding of the current data in the neural network. With the wide application of time series models based on the RNN, greater precision of prediction and the long short-term memory recurrent network (LSTM-RNN) \cite{azzouni2017long} has been proposed. LSTM-RNN not only retains the timing characteristics of the RNN structure, but also has the memory function for time series.

The systematic mapping presented in this paper investigates the following questions: (i) the vehicle and dates of publication of the studies; (ii) the contexts of problems that have been addressed in LSTM networks related to Time Series; (iii) the LSTM models that are being used to treat problems in Time Series; (iv) effectiveness of LSTM models in predictions; (v) which are the hyperparameter configurations of the networks; (vi) which air quality characteristics are being used the most; (vii) what other neural network methods are being used in conjunction with the LSTM. This mapping was structured in 7 research questions, where 155 studies were selected and analyzed, 57 of which were approved by reading the abstract and keywords, and from these, 12 had the data extracted, according to the systematic mapping method used.

This article is organized as follows. Section 2 presents a theoretical foundation about the research. Section 3 the research protocol used. Section 4 the main results. Section 5 a discussion of the results. Section 6 concludes the work.

\section{Background} \label{sec:firstpage}
\label{sec-lstm}
In this section, to aid in the decision-making of research questions, which is the object of systematic mapping, and define the scope of our investigation, we discuss some of the key concepts in the research areas studied (LSTM).

\subsection{LSTM}
\label{subsec-lstm}
The original LSTM idea was initially proposed by \cite{hochreiter1997long} and since then different models have been proposed to improve the performance of LSTM \cite{cho2014learning}, \cite{li2019ea}, \cite{karevan2020transductive}, \cite{hu2020time}. The LSTM uses a locking mechanism to control the information that must be maintained over time, the duration that must be maintained and the time that can be read through the memory cell.

The LSTM recurrent neural network consists of a module with memory cells that can learn data characteristics in the time domain. It has been widely used in many fields due to its improved processing performance for time series data. The memory module in the LSTM recurrent neural network contains three multiplication units: an input gate, a forgetting gate and an output gate. These gates control the input, update and output of information, respectively, so that the network has a certain memory function. On the other hand, the network has more learning parameters due to these gates.

\section{Research Protocol}
This systematic mapping study was performed following the guidelines proposed by the systematic literature reviews  that involve three main phases \cite{keele2007guidelines}. The Planning phase refers to the preview activities and aims at establishing a review protocol defining the research questions, inclusion and exclusion criteria, study sources, search string, mapping procedures and the control papers. These control papers are works that closely match with the objectives of the study and the inclusion and exclusion criteria used to define the selected papers. Thus, the control papers must be part of the selected works.

The Conducting phase consists of searching, selecting and analyzing the studies, in order to extract and synthesize their data. This phase executes the plan defined in the Planning phase. The last is the Report phase, which is characterized by the writing of the results to disseminate them. The main results from the first phase are presented below.

\subsection{Research Questions}
This mapping aims at answering the following research questions:

\begin{itemize}
    \item[\textbf{RQ1.}] When and where have the studies been published?
    \item[\textbf{RQ2.}] Which contexts have been most approached in LSTM networks related to Time Series?
    \item[\textbf{RQ3.}] Which LSTM models are being used to treat time series problems?
    \item[\textbf{RQ4.}] What hyperparameters have been used in LSTM networks?
    \item[\textbf{RQ5.}] What air quality characteristics have been most approached?
    \item[\textbf{RQ6.}] What other neural network methods are being used in conjunction with LSTM?
    \item[\textbf{RQ7.}] Are LSTM models effective in predicting Time Series?
\end{itemize}

\subsection{Inclusion and Exclusion Criteria}
Only 1 inclusion criteria (IC) and 8 exclusion criteria (EC) were created. The inclusion criterion is: (IC1) The study uses LSTM in data time series on air quality. The exclusion criteria are: (EC1) The study does not have an abstract; (EC2) The study is just published as an abstract; (EC3) The study is not written in English; (EC4) The study is an older version of another study already considered; (EC5) The publication is a tutorial, events annals, lecture log or secondary study; (EC6) It was not possible to access the study; (EC7) The study does not satisfy the inclusion criteria; (EC8) Does not use data about PM$_{2.5}$ or PM$_{10}$.

\subsection{Sources}
The search was applied to 7 electronic databases. However, it was not possible to export data from some databases used, so the results of the research of only two of them were used, which are:

\begin{itemize}
    \item \textbf{Scopus} (http://www.scopus.com)
    \item \textbf{IEEE Xplore} (https://ieeexplore.ieee.org)
\end{itemize}

\subsection{Keywords and Search String}
The search string used in this study considered three areas: the LSTM neural network, Time Series and Air Pollution (see Table \ref{tabela:1}), and was applied to three metadata fields (title, abstract and keywords). The string has been refined about 7 times to fit the proposed study topic.

\begin{table}
\centering
\caption{ Keywords and Search String}
\begin{tabular}{ |c|c| } 

 \hline
 \textbf{Area} & \textbf{Keywords}  \\ 
 \hline
 LSTM & "LSTM", "\textit{long short-term memory}", "\textit{prediction}", "\textit{prescription}" \\ 
 \hline
 Time Series & "\textit{Time Series}", "\textit{forecasting}", "\textit{forecast}"\\
 \hline
 Air Pollution & "\textit{air}", "\textit{pollution}" \\
 \hline
 \multicolumn{2}{|p{12cm}|}{\textbf{Search String}: (("LSTM"\ OR "long short-term memory") AND  "Time Series"\ AND  ("forecast*"\ OR "prediction"\ OR "prescription") AND  ("air"\ OR "pollution"))}\\
 \hline
\end{tabular}
\label{tabela:1}
\end{table}

\subsection{Data Storage}
The studies returned in the searching phase were cataloged appropriately. This helped to classify and analyze the studies for the next phase more clearly and intelligently.

\subsection{Assessment}
The mapping protocol was tested to verify its feasibility and adequacy, based on a pre-selected set of studies considered relevant for the investigation. The review process was conducted by the first author from this paper, who carried out its validation. About 37\% of the studies were analyzed.

\section{Conduction the Mapping}
In this section, we discuss the main steps of this research, namely: search and selection, and synthesis and data analysis.

\subsection{Search and Selection}

Only articles from 2013 onwards were considered because as the theme is recent, no publications were found before this date. As a result, 155 publications were returned, of which 109 from \textbf{Scopus}, and 46 from \textbf{IEEE Xplore}.

A selection process for the returned publications was applied, which was divided into 3 stages. In the first, duplicates are eliminated based on examining the title and abstract, and the number of publications has been reduced to 123 (20.64\% reduction). In the second step, the inclusion and exclusion criteria were imposed considering the title and abstract. 65 publications (52.84\%) were eliminated. Some papers had a slightly confused title and abstract and made it difficult to apply the inclusion and exclusion criteria. This made it necessary to read the introduction of these studies to apply the criteria.

Finally, in the third step, the inclusion and exclusion criteria were applied considering the entire text, reducing the number of publications from 58 to 12 (79.31 \%). Unfortunately, many publications with excellent abstracts were eliminated because it is not possible to have access to the entire text. Attempts have been made to contact some of the authors of these excluded papers, but there was still no feedback.

From these steps, we selected 12 studies that were considered relevant, 9 + 3 control papers, for the next phase: Data Analysis. Table \ref{tabela:2} shows the steps and results of the selection process. The selection process resulted in a reduction of about 92\%, 12 out of 155. Table \ref{tabela:3} lists these 12 studies that were considered relevant.

\begin{table}
\centering
\caption{ Selection Process Steps Results}
\begin{tabular}{ |p{1cm}|p{2cm}|p{2cm}|p{2cm}|p{2cm}|p{2cm}| } 
 \hline
 \textbf{Step} & \textbf{Criteria} & \textbf{Analyzed Content} & \textbf{Initial N. of Studies} & \textbf{Final N. of Studies} & \textbf{Reduction (\%)}  \\ 
 \hline
 1ª & Eliminating Duplication & Title and Abstract & 155 & 123 & 20,64\% \\
 \hline
 2ª & IC1, EC3, EC5, EC6, EC7 & Title and Abstract & 123 & 58 & 52,84\% \\
 \hline
 3ª & IC1, EC8 & Entire Text & 58 & 12 & 79,31\% \\
 \hline
\end{tabular}
\label{tabela:2}
\end{table}

\begin{table}
\centering
\caption{ Selected Papers}
\begin{tabular}{ |p{1cm}|p{4cm}|p{1cm}|p{7cm}| } 
 \hline
 \textbf{ID} & \textbf{Reference} & \textbf{ID} & \textbf{Reference} \\ 
 \hline
 [1] & \cite{singh2019deepair} & [2] & \cite{zhao2019long} \\
 \hline
 [3] & \cite{riekstin2018time} & [4] & \cite{li2020hybrid} \\
 \hline
 [5] & \cite{lu2020new} & [6] & \cite{qin2019novel} \\
 \hline
 [7] & \cite{wu2019novel} & [8] & \cite{liu2019self} \\
 \hline
 [9] & \cite{liu2020air} & [10] & \cite{thaweephol2019long} \\
 \hline
 [11] & \cite{park2017pm10} & [12] & \cite{li2020urban} \\
 \hline
\end{tabular}
\label{tabela:3}
\end{table}

\subsection{Synthesis and Data Analysis}
In this section, answers to research questions are presented.

\textbf{RQ1. When and where have the studies been published?} -- The selected publications were published between 2017 and 2020, being that in 2019 there is a higher occupancy rate (50\% of publications). In relation to type of publication vehicle, 75\% were published in Journals and 25\% in Conferences, highlighting the IEEE Access with 25\% of publications.

\begin{figure}
\centering
\includegraphics[width=0.65\textwidth]{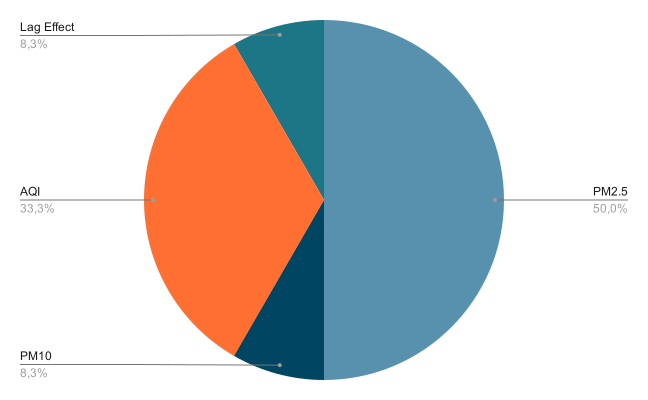}
\caption{Predictions made by the studies}
\label{aqi}
\end{figure}

\textbf{RQ2. Which contexts have been most approached in LSTM networks related to Time Series?} -- The main prediction problems were in relation to PM$_{2.5}$, Figure \ref{aqi} shows the relationships of the predictions made in the studies. Standing out \cite{lu2020new} for making predictions of the lag effect of pollutants on respiratory diseases, and \cite{liu2019self} for using predictions to treat haze-fog problems.

\textbf{RQ3. Which LSTM models are being used to treat time series problems?} -- Among the models used in the publications, 67\% were the standard LSTM and 33\% were a new version of the LSTM. In this one we highlight one of the control papers, \cite{zhao2019long} by LSTM-FC, and also \cite{li2020hybrid} by CNN-LSTM and \cite{li2020urban} by AC-LSTM which is a variant of CNN-LSTM with a layer of attention mechanisms (attention-based layer).

\textbf{RQ4. What hyperparameters have been used in LSTM networks?} -- The hyperparameters used vary widely in each publication, among them the number of LSTM layers varied from 1 to 64, the nodes varied from 16 to 800 for each layer. Batch size was 2 to 108, epochs from 20 to 639. About the activation function, 42 \% opted for sigmoid, 25\% opted for ReLu and 33\% did not specify. 16\% used optimizer Adam, and 16\% also used RMSProp, highlighting \cite{park2017pm10} who used Adagrad in addition to RMSProp and compared the performance between the two, in which it can be seen that RMSProp performed better.

\textbf{RQ5. What air quality characteristics have been most approached?} -- Among the pollutants approached, PM $_{2.5}$ had a greater focus with a presence in 83\% of the articles, followed by PM10 with 58\%. In addition, 50\% of the publications used other pollutants related to the Air Quality Index (AQI), such as \cite{thaweephol2019long} which is able to correlate the growth of PM$_{2.5}$ with the other pollutants.

\begin{figure}
\centering
\includegraphics[width=0.65\textwidth]{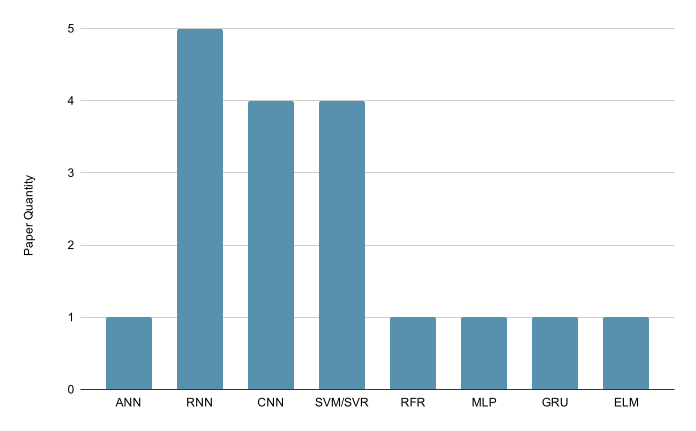}
\caption{Models used with LSTM. (GRU - Gated recurrent unit, and ELM - Extreme learning machine)}
\label{redes}
\end{figure}

\textbf{RQ6. What other neural network methods are being used in conjunction with LSTM?} -- CNN, RNN and SVR (support vector regression) are the most outstanding in relation to the networks that are being used together or for comparisons. Figure \ref{redes} shows all the models used. The paper \cite{li2020urban} stands out for making comparisons with SVR, RFR (random forest regression), MLP (multilayer perceptron), RNN, LSTM, CNN-LSTM and presenting superior results to all of them.

\textbf{RQ7. Are LSTM models effective in predicting Time Series?} -- All publications used metrics of moving averages of the error comparing the proposed model with other models. The metrics used were Mean Absolute Error (MAE), Mean Absolute Percentage Error (MAPE), Root Mean Squared Error (RMSE), and Pearson Correlation. The variations in error reduction goes from 7.45\% to 500\% as in the study \cite{park2017pm10}. Another notable is \cite{qin2019novel} where Pearson correlation reaches 0.97, which is very close to the perfect one where 1 would be the perfect correlation between the real and the predicted value.

\section{Final Considerations}
This paper presented a systematic mapping that investigates the use of LSTM neural networks in time series data on air quality. 155 studies were investigated and 12 were selected for further analysis. Despite the large number of publications returned from the search engines, it is possible that there are still publications that were not identified. A future work here is to enlarge the number of studies to get more accurate data or merge the data obtained by this research with others studies.

The contributions of this work are on showing and comparing the LSTM models used on time series prediction. In this context, we highlight the following conclusions: (i) A lot of studies in this area started to emerge after 2017, this shows a growing concern about air quality; (ii) Some studies already show that it is possible to create LSTM networks with unique characteristics for specific themes, thus improving their performance when compared to the standard model, as well as the creation of hybrid models using other neural networks together from LSTM as seen in \cite{zhao2019long}, \cite{li2020hybrid} e \cite{li2020urban}; (iii) Unfortunately, not all studies specified their network hyperparameters, but it is still possible to notice a preference in the activation function, sigmoid and ReLu, and in the optimizer, Adam and RMSProp; (iv) The effectiveness of LSTM models for predicting time series is remarkable in comparison with the statistical models in the literature and some other models of neural networks. Most of the studies obtained gains above 50\% in the reduction of errors of the metrics MAE, MAPE and RMSE.

It is undeniable that concern with air quality is one of the areas that has grown the most in this area in recent years. And, as we can see, the use of LSTM models is, in fact, positively influencing the prediction of the main air pollutants. With the growth of IoT technologies and the advancement of artificial intelligence, it is promising to say that these improvements tend to become more and more optimized and with more accurate tools it is possible to improve the quality of life in relation to air quality.

\bibliographystyle{sbc}
\bibliography{sbc-template}

\end{document}